\documentclass[manuscript]{copernicus}

\begin{document}

\title{OpenEM: Large-scale multi-structural 3D datasets for electromagnetic methods}


\Author[1,2]{Shuang}{Wang} 
\Author[1,2][wxb@cdut.edu.cn]{Xuben}{Wang}
\Author[3]{Fei}{Deng}
\Author[1,4]{Peifan}{Jiang}
\Author[5]{Jian}{Chen}
\Author[5]{Gianluca}{Fiandaca}

\affil[1]{Key Laboratory of Earth Exploration and Information Techniques of Ministry of Education, Chengdu University of Technology, Chengdu, China.}
\affil[2]{College of Geophysics, Chengdu University of Technology, Chengdu, China.}
\affil[3]{College of Computer Science and Cyber Security, Chengdu University of Technology, Chengdu, China.}
\affil[4]{Faculty of Geo-Information Science and Earth Observation, University of Twente, Enschede, Netherlands}
\affil[5]{The EEM Team for Hydro and eXploration, Department of Earth Sciences “Ardito Desio”, University of Milano, Milano, Italy}




\runningtitle{TEXT}

\runningauthor{TEXT}

\received{}
\pubdiscuss{} 
\revised{}
\accepted{}
\published{}


\firstpage{1}

\maketitle

\nolinenumbers

\begin{abstract}
Electromagnetic (EM) methods, owing to their efficiency and non-invasive nature, have become one of the most widely used techniques in geological exploration. Nevertheless, data processing for these methods remains highly time-consuming and labor-intensive. With the remarkable success of deep learning, applying such techniques to EM methods has emerged as a promising research direction to overcome the limitations of conventional approaches. The effectiveness of deep learning methods depends heavily on the quality of datasets, which directly influences model performance and generalization ability. Existing application studies often construct datasets from random one-dimensional or structurally simple three-dimensional (3D) models, which fail to represent the complexity of real geological environments. Furthermore, the absence of standardized, publicly available 3D geoelectric datasets continues to hinder progress in deep learning–based EM exploration. To address these limitations, we present OpenEM, a large-scale, multi-structural 3D geoelectric dataset that encompasses a broad range of geologically plausible subsurface structures. OpenEM consists of nine categories of geoelectric models, spanning from simple configurations with anomalous bodies in half-space to more complex structures such as flat layers, folded layers, flat faults, curved faults and their corresponding variants with anomalous bodies. In addition, we provide a 3D model generator that enables fully controllable 3D model construction, allowing flexible and extensible augmentation of OpenEM. OpenEM provides a unified, comprehensive, and large-scale dataset for common EM exploration systems to accelerate the application of deep learning in electromagnetic methods. The complete dataset and 3D model generator is publicly available at https://doi.org/10.5281/zenodo.17141981 \cite[]{OpenEM}.
\end{abstract}


\introduction  
Electromagnetic (EM) methods are widely employed in geophysical exploration and remain a central focus of research in the geological exploration industry. A variety of EM systems have been developed, including ground-based, airborne, semi-airborne, time-domain and frequency-domain electromagnetic systems. These systems have been extensively applied to geological hazard assessment \cite[]{damhuis2020identification,malehmir2016near}, groundwater detection \cite[]{ball2020high,minsley2021airborne}, mineral resource exploration \cite[]{kone2021geophysical,okada2021historical}, and geological mapping \cite[]{dzikunoo2020new,wong2020interpretation}.

The extraction of geoelectric structural information from EM data primarily involves two processes: data processing and inversion \cite[]{wu2022deep}. Apart from correcting for system response effects, the core task of data processing is denoising. Conventional denoising methods generally depend on empirically chosen parameters, which place heavy demands on the operator’s expertise \cite[]{wu2019removal,wu2020removal}. Inversion provides direct insights into the geoelectric structure, however, conventional approaches require iterative corrections through forward modeling, making the process computationally intensive \cite[]{siemon2009laterally,vallee2009inversion,vignoli2015sharp,christensen2017voxel}. The computational cost increases rapidly with the number of observations, making it unbearable when dealing with large volumes of data \cite[]{wu2023fast}.

With the rapid development of deep learning, replacing conventional approaches with deep learning techniques to overcome their inherent limitations has gained significant attention and led to numerous practical applications \cite[]{huang2025data,wu2025frontiers,wu2025characterizing}. For instance, deep learning has been applied to accelerate forward modeling and mitigate the high computational cost of conventional methods \cite[]{bording2021machine,qu2025deep,asif2021effect,wu2023deep}, to process acquired data and reduce dependence on empirical expertise \cite[]{asif2025comparative,wu2021noising,asif2022automated,li2024gtcn,wang2025dremnet}, and to perform inversion for efficiently handling large datasets and rapidly obtain subsurface electrical structures \cite[]{chen2022transient,wu2021convolutional,chen2025rapid,wu2022instantaneous,wu2024physics,wu2025variational}. The effectiveness of deep learning methods largely depends on the quality of the dataset, which directly influences both model performance and generalization capability \cite[]{wang2025interpretable}. Existing approaches, however, either rely on subsurface resistivity models derived from field data or generate large sets of resistivity models through random or pseudo-random procedures. Training on field-derived resistivity models allows deep learning algorithms to capture geological priors of the target survey area \cite[]{bai2020quasi}, but restricts their applicability to that region, limiting generalization to geologically distinct environments. In contrast, randomly generated resistivity models assign resistivity values to each layer based on probability distributions, often resulting in geologically implausible structures \cite[]{asif2023dl}. Consequently, models trained on such dataset struggle to achieve optimal performance.

Therefore, to enhance the performance of deep learning models, improve their ability to process field data, and promote the development of deep learning techniques tailored for electromagnetic methods, a unified, comprehensive, and large-scale resistivity model dataset is urgently required \cite[]{bergen2019machine,reichstein2019deep}. \cite{asif2023dl} introduced DL-RMD, a large one-dimensional resistivity dataset that has been widely used in the community for tasks such as denoising \cite[]{liu2024multi} and inversion \cite[]{zhu2025deep}. However, the real world is inherently three-dimensional (3D), characterized by spatially continuous variations, whereas one-dimensional resistivity models cannot capture complex geological structures such as faults and intrusions \cite[]{akingboye2025electrical,teklesenbet2012multidimensional,melo2018integrated}. Moreover, due to 3D effects and the interaction between subsurface anomalies and measurement errors, one-dimensional inversion may produce significantly different results at different observation points. Such inconsistencies often lead to poor lateral continuity in inversion profiles and may even generate spurious anomalies \cite[]{yang2012three,oldenburg20203d}. To address the limitations of one-dimensional models, researchers have begun exploring 3D inversion with deep learning. Nevertheless, the datasets employed thus far remain relatively simple, typically consisting of half-spaces with anomalies \cite[]{tang2024fast,zhang20253d,zhao2024three}. These simplified models fail to adequately represent the wide range of complex geological structures observed in reality, thereby limiting their applicability to field data processing. Furthermore, the lack of a unified and comprehensive 3D resistivity model dataset continues to present a major obstacle to the broader application of deep learning in electromagnetic methods.

To address the limitations and provide a unified, comprehensive, large-scale dataset for advancing the application of deep learning in electromagnetic methods—as well as to supply reliable dataset for field data processing—we present OpenEM, a large-scale, multi-structural, 3D geoelectric model dataset that encompasses a wide range of geologically plausible subsurface structures. OpenEM contains nine categories of geoelectric models with varying levels of complexity, including half-space models with anomalous bodies, flat layers, folded layers, flat faults, curved faults, and their corresponding variants containing anomalous bodies. The resistivity values range from 1 to 2000 Ω·m, and the models consist of 3 to 7 layers. In models with anomalies, 1 to 5 anomalous bodies of both regular and irregular shapes are incorporated to enhance dataset diversity and realism. We further provide OpenEM with an accompanying 3D model generator that enables fully controllable 3D model construction, allowing users to customize structural characteristics of the generated models, including resistivity magnitudes, fault geometries and locations, as well as the size, shape and spatial distribution of anomalous bodies. 

OpenEM is compatible with widely used electromagnetic exploration systems, including ground-based, airborne, semi-airborne, time-domain, and frequency-domain electromagnetic systems. The dataset can be employed to construct surrogate models for well-established processes and supports computationally intensive tasks such as denoising, forward modeling, and inversion. Moreover, OpenEM facilitates research in related areas, including complexity analysis, uncertainty quantification, and generalization studies.
\begin{figure*}[t]
	\includegraphics[width=12cm]{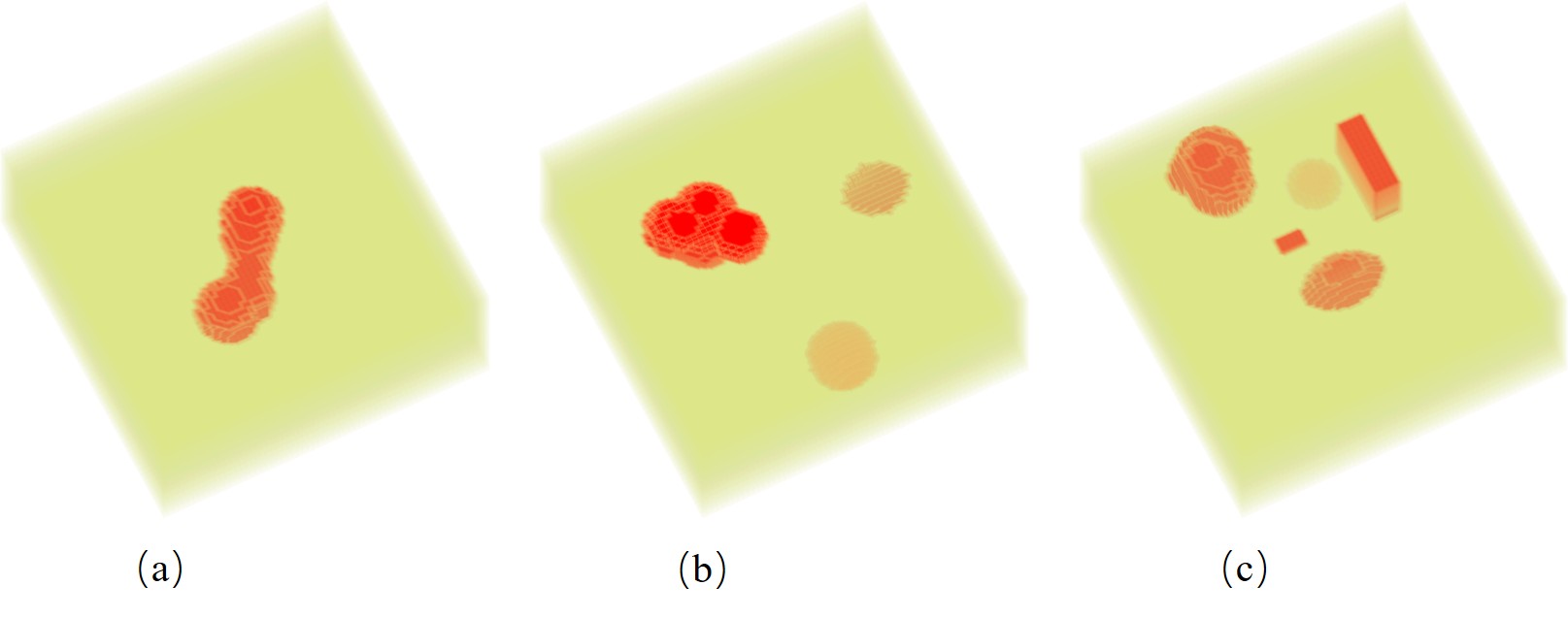}
	\caption{Examples of half-space models: (a) contains one irregular anomaly, (b) contains three anomalies, (c) contains five anomalies.\label{1}}
\end{figure*}
\section{Methodology}
Geological processes rarely produce random structures, and subsurface resistivity variations are similarly non-random, instead, they typically exhibit spatial correlations, with geological formations largely composed of sedimentary layers. Consequently, randomly generated models are inadequate for realistically representing geological conditions and often fail to satisfy practical requirements. To provide a unified, comprehensive, large-scale dataset that supports the application of deep learning in electromagnetic methods and provides reliable resources for field data processing, we constructed both simple half-space models with anomalies and more complex multi-structural models.
\subsection{Half-space models with anomalies}
For the half-space models, background resistivity is set between 100 and 1000 Ω·m in increments of 100 Ω·m, while anomaly resistivity ranges from 1 to 2000 Ω·m. Anomalies are categorized into regular anomalies and irregular anomalies. Regular anomalies include quadrangular prisms, triangular prisms, spheres, and ellipsoids. The number of anomalies is randomly assigned between 1 and 5. Fig. ~\ref{1} shows three examples of half-space models containing 1, 3 and 5 anomalies, illustrating both regular and irregular types.

\subsection{Multi-structure models}
To generate geologically plausible models, we employed von Kármán covariance functions \cite[]{moller2001rapid} to create the initial layered geological structures.
\begin{equation}
r = A^2 C \left( \frac{z}{L} \right)^\nu K_\nu \left( \frac{z}{L} \right)
\end{equation}
Where, $A$ represents the amplitude of the logarithmic resistivity, $C$ is the scale factor, $z$ denotes the spatial (vertical) distance, $L$ is the maximum correlation length considered, and $K$ is the modified Bessel function of the second kind with order $v$. 

The generated initial layered model is presented in Fig. ~\ref{2}a. Building on this model, geological structures such as faults and folds are iteratively incorporated. Letting the initial layered model be denoted as $r_0$, faults are introduced through iterative updates according to the following mathematical formulation:
\begin{figure*}[t]
	\includegraphics[width=12cm]{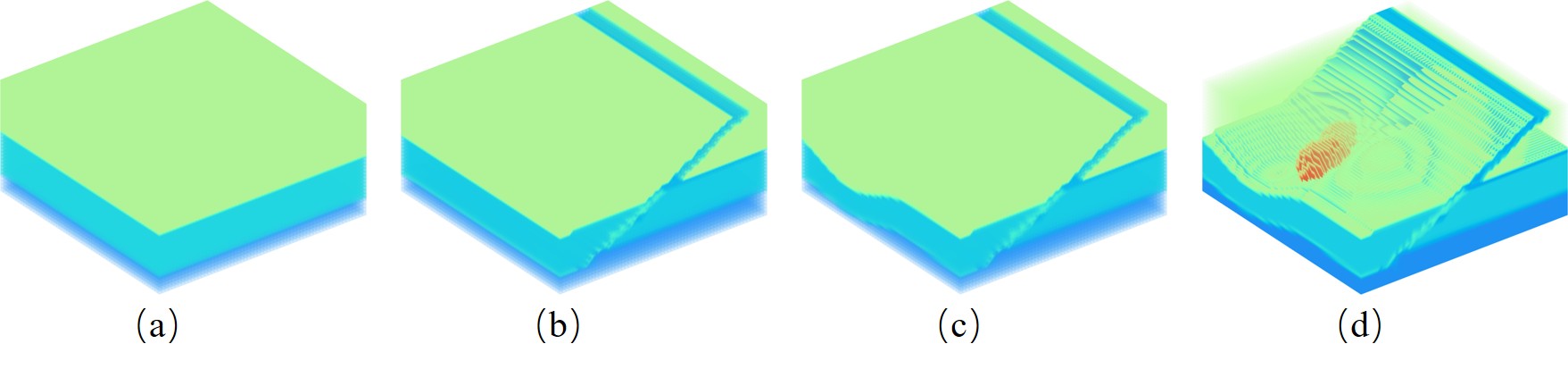}
	\caption{Model construction process: (a) initial layered model, (b) model with added faults, (c) model with added folded strata, and (d) model with embedded anomalies.\label{2}}
\end{figure*}
\begin{equation}
r_i(x, y, z) = 
\begin{cases} 
	r_0(a_i \sin(2\pi k_i x) + s_i, a_i \sin(2\pi k_i y) + s_i, z + s'_i), & z \geq f_i(x,y) \\
	r_{i-1}(x, y, z), & z < f_i(x,y) 
\end{cases}; \quad i > 0
\end{equation}
Where, $s_i$, $s_i^\prime$, $k_i$, and $a_i$ are the random variables at the $i$-th iteration. $f_i(x,y)$ denotes a random curve used for fault simulation.
\begin{equation}
f_{i}(x,y) = c_ix + d_iy + A_1 \sin(\omega_1 x + \phi_1) + A_2 \cos(\omega_2 y + \phi_2) + e_i
\end{equation}
Where $c_i$, $d_i$, $A_1$, $A_2$, $\omega_1$, $\phi_1$, $\omega_2$, $\phi_2$, and $e_i$ denote the random variables at the $i$-th iteration, and the model after incorporating faults is illustrated in Fig. ~\ref{2}b. Once the fault structures are established, folds are subsequently introduced into the model, which are simulated using the following mathematical formulation:
\begin{equation}
r_i(x, y, z) = r_{i-1}(x, a_i \sin(2\pi k_i x), a_i \sin(2\pi k_i x)); \quad i > 0
\end{equation}
The model after incorporating folds is illustrated in Fig. ~\ref{2}c.
\begin{figure*}[t]
	\includegraphics[width=14cm]{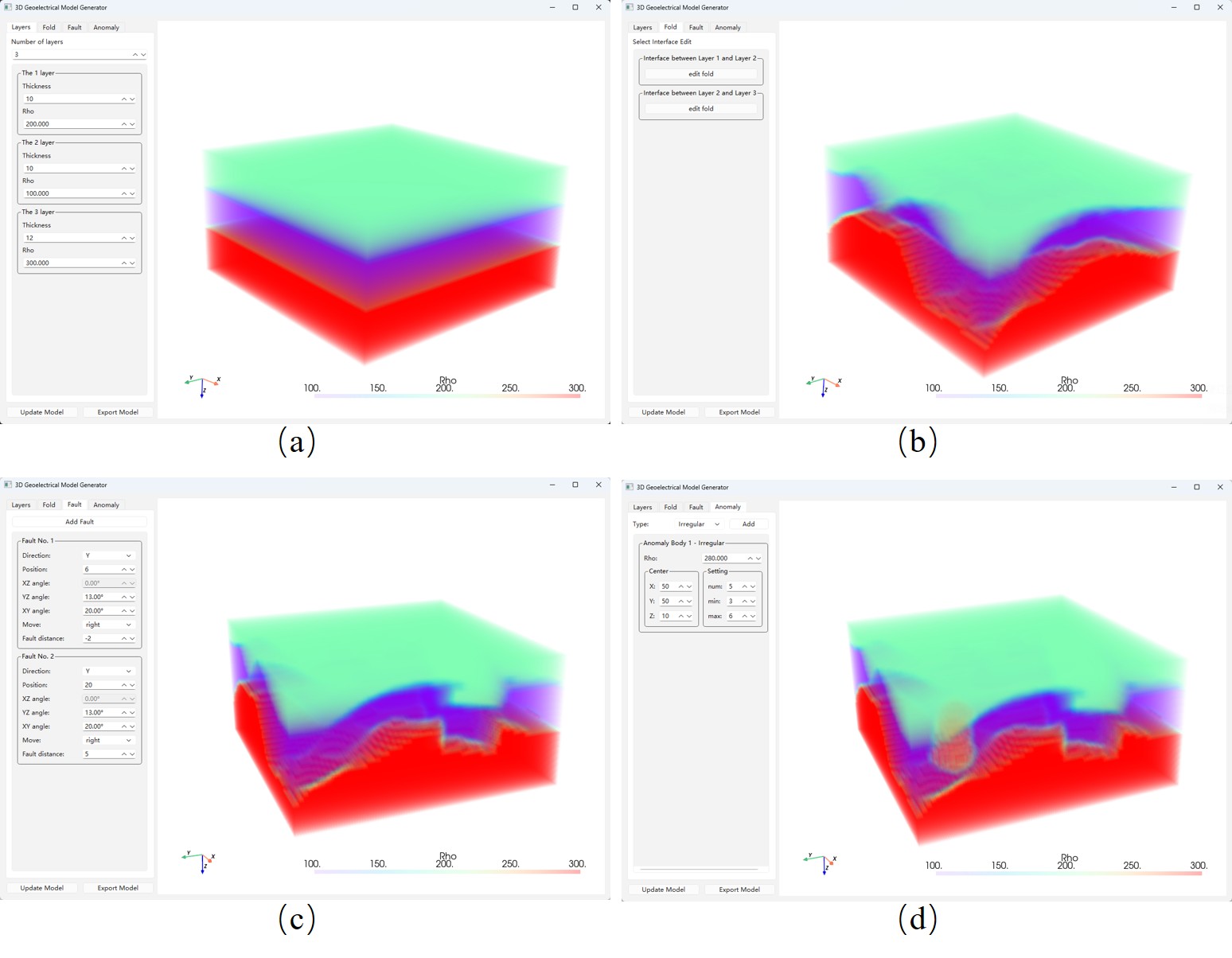}
	\caption{3D model generator, (a) generation of a layered model, (b) model with curved stratification, (c) model incorporating faults and (d) model with embedded anomalous bodies.\label{3}}
\end{figure*}

Through the construction process described above, layered geological models containing faults and folds are generated. Alternatively, one or two of these steps can be applied independently to create different types of geological models. To simulate resistivity variations induced by local anomalies, 1 to 5 anomalies are subsequently introduced into the model. Their scales are consistent with those used in the half-space models and are classified as either irregular or regular. Regular anomalies primarily include quadrangular prisms, triangular prisms, spheres, and ellipsoids. The number of anomalies is randomly assigned between 1 and 5. The model with anomalies added is illustrated in Fig. ~\ref{2}d.
\subsection{3D model generator}
\begin{figure*}[t]
	\includegraphics[width=14cm]{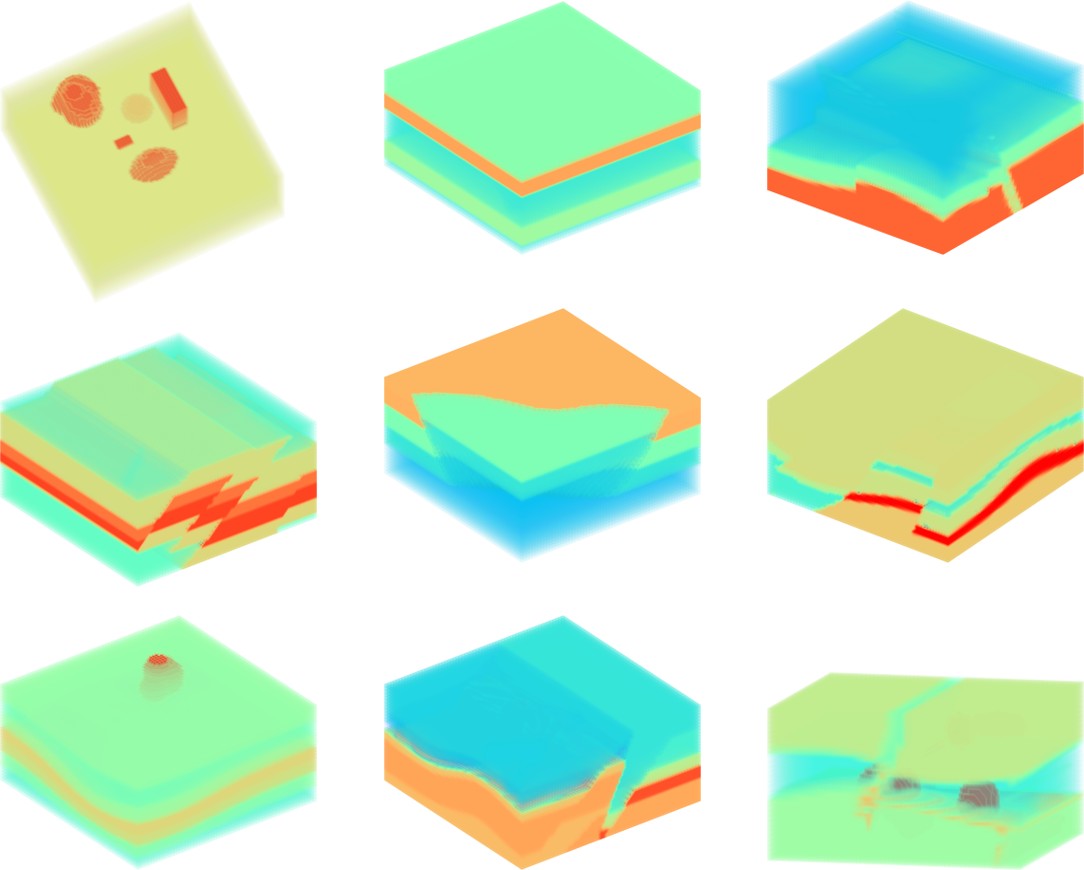}
	\caption{Examples of OpenEM models, illustrating the nine types included in OpenEM.\label{4}}
\end{figure*}
Models constructed via mathematical transformations are capable of representing most geological scenarios, however, certain extreme geological conditions may still fall outside the scope of such models, for example when the required resistivity range exceeds 2000 Ω·m. To complement these limitations and to provide broader model coverage, we introduce a fully controllable 3D model generator that enables comprehensive customization of geoelectrical models. As illustrated in Fig. ~\ref{3}a, the desired stratigraphic units and their resistivity values can be specified, after which the stratigraphic surfaces can be interactively edited to form undulating layers. In addition, faults can be generated in a fully customized manner, as shown in Fig. ~\ref{3}c, and a variety of regular and irregular anomalous bodies can be incorporated, as depicted in Fig. ~\ref{3}d.
\section{OpenEM}
OpenEM consists of nine types of geological models, as illustrated in Fig. ~\ref{4}. These models range from simple to complex and include half-space models with anomalies, layered models, layered models with faults, folded models, folded models with faults, and their respective variants containing anomalies. Each type comprises approximately 120,000 models, yielding a total of about 1.08 million models in OpenEM. Collectively, these models capture the majority of geological scenarios encountered in practice. The key features of OpenEM are summarized in Table 1.

\begin{table*}[h]
\caption{OpenEM model information}
\begin{tabular}{ccccc}
\tophline
        & Number   & Resistivity    & Number of layer & Number of anomaly \\
\middlehline
OpenEM  & 1,080,000   & 1-2000    & 3-7 & 1-5  \\
\bottomhline
\end{tabular}
\belowtable{} 
\end{table*}

\begin{figure}[t]
	\includegraphics[width=8.3cm]{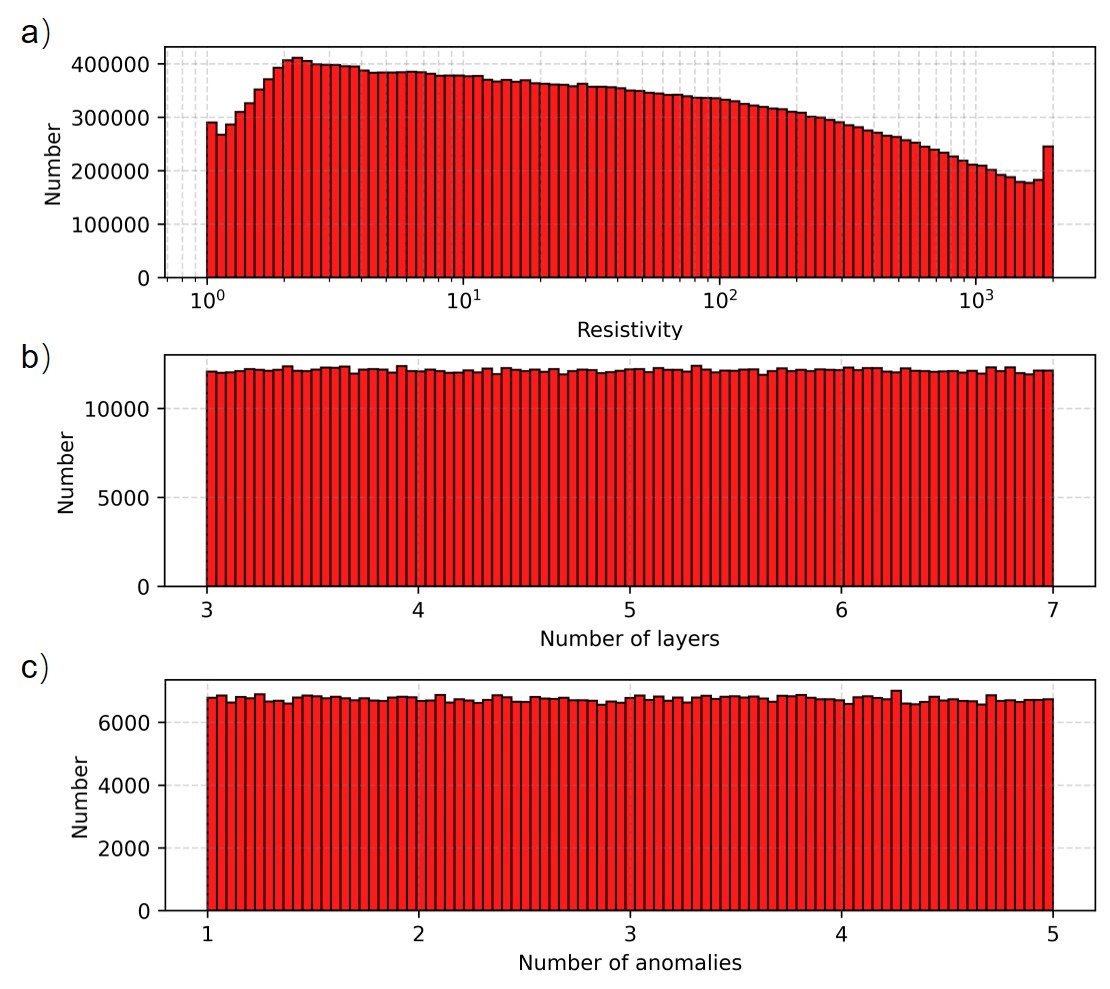}
	\caption{Statistical information of OpenEM. (a) Resistivity distribution, (b) Number of layer distribution, (c) number of anomaly distribution.\label{5}}
\end{figure}

The statistical characteristics of OpenEM are presented in Fig. ~\ref{5}. Fig. ~\ref{5}a depicts the overall resistivity distribution, which is approximately uniform but slightly skewed toward lower resistivity values. This distribution is favorable for electromagnetic methods, as their resolution is generally higher in low-resistivity regions \cite[]{jorgensen2005contributions}. Fig. ~\ref{5}b shows the statistics of the number of layers in layered models, where the layer count is uniformly distributed between 3 and 7. Fig. ~\ref{5}c illustrates the statistics of anomaly counts, indicating that the number of anomalies ranges from 1 to 5 and follows a uniform distribution.
\section{Forward}
\begin{figure*}[t]
	\includegraphics[width=14cm]{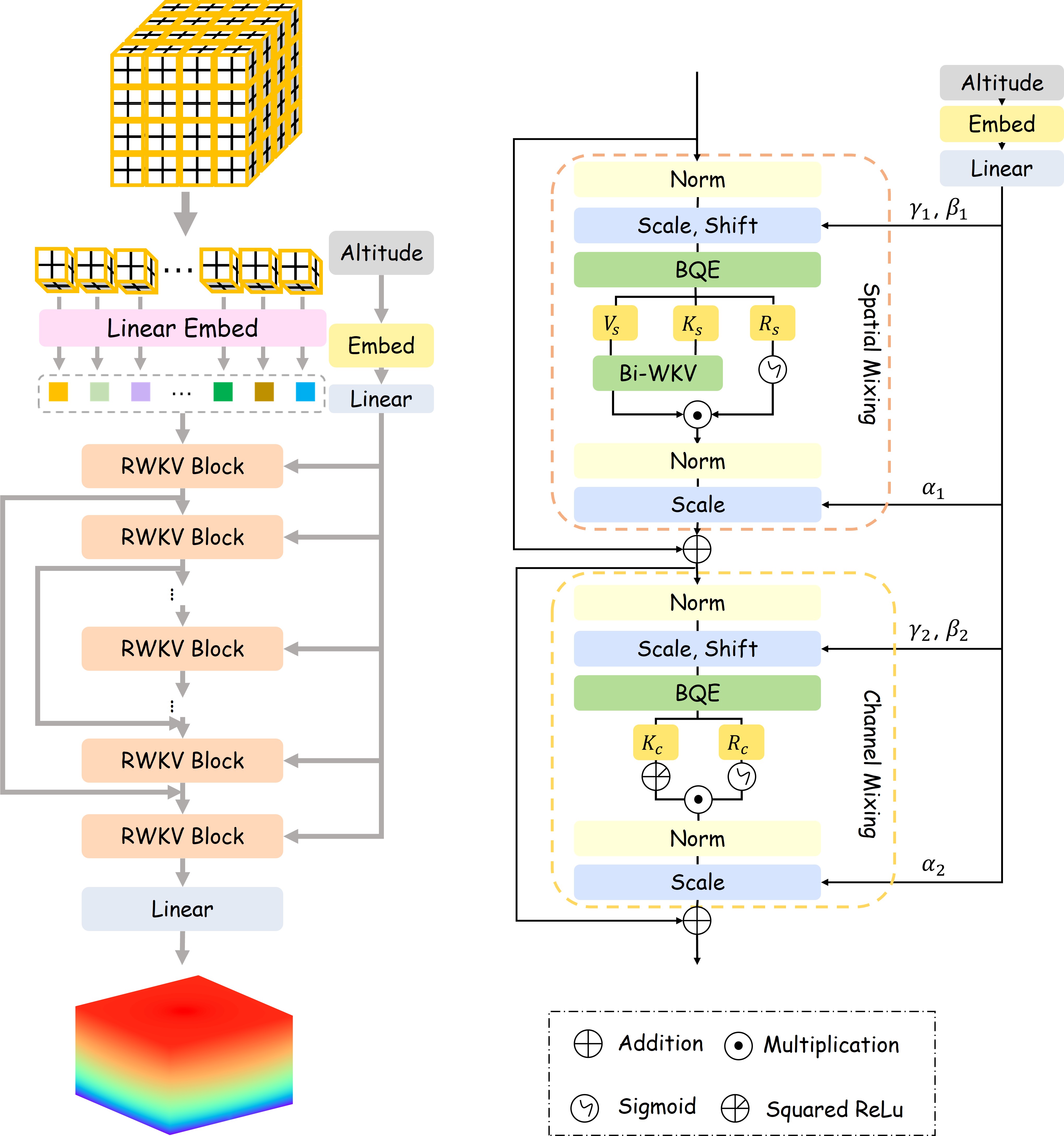}
	\caption{Forward modeling network architecture, modified from \cite{wang2026rapid}. The overall structure shown on the left and detailed module information on the right.\label{6}}
\end{figure*}
When applying OpenEM in practical applications, it is necessary to perform forward modeling on the models to generate usable labels. However, forward modeling of 3D models is computationally expensive, which presents challenges for practical applications. Therefore, the method proposed by \cite{wang2026rapid} can be adopted to perform rapid forward modeling for OpenEM.
\subsection{Network architecture}
The overall architecture of the network is shown in Fig. ~\ref{6}. The network is built based on RWKV, taking a 3D geoelectric model as input and outputting the corresponding forward modeling data. The entire network consists of stacked RWKV blocks. After 3D embedding, the geoelectric model is processed by a series of RWKV blocks, ultimately outputting the forward modeling results. To address the variations caused by altitude, the RWKV blocks process the transceiver altitude separately. By linearly embedding the altitude to modulate the RWKV blocks, the network is made sensitive to the transceiver altitude.
\subsection{Train}
A total of 10,000 models were randomly selected from OpenEM for forward modeling, uniformly sampled across nine types of geological models. The dataset was then divided into training, validation, and test subsets in an 8:1:1 ratio.

The forward modeling was conducted according to the specifications of the AeroTEM IV system \cite[]{bedrosian2014airborne}, including the transmitter current and receiver configuration. The corresponding responses were computed using the SimPEG forward modeling framework \cite[]{cockett2015simpeg}. The computational domain is defined with a length and width of 640 m and a depth of 320 m, while the transceiver altitude is varied between 25 and 100 m, representing typical flight conditions. The observation system was uniformly distributed, consisting of 32 survey lines, each with 32 measurement points, and 32 selected time samples, resulting in output responses of size 32 × 32 × 32.

The network was trained for 200 epochs on a single NVIDIA A6000 GPU using the AdamW optimizer with a learning rate of 0.0001.
\subsection{Test}
\begin{figure*}[t]
	\includegraphics[width=18cm]{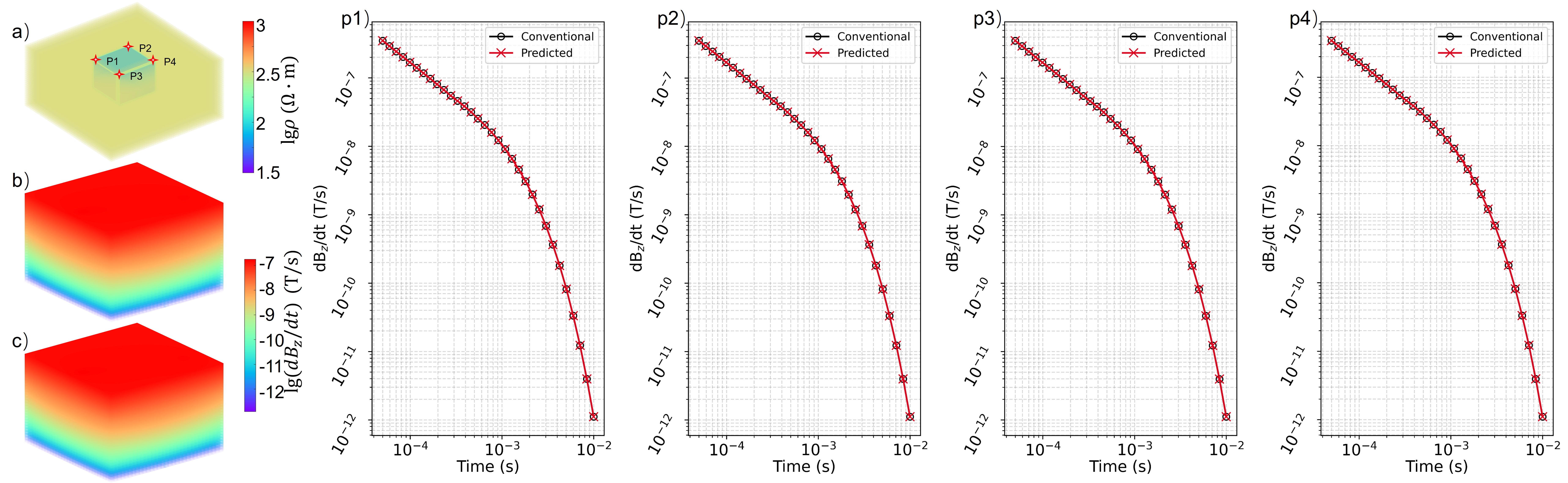}
	\caption{Simple model test: (a) geoelectric model, (b) network-predicted results, (c) forward modeling results using the conventional method, and (P1)–(P4) correspond to the responses at the four points indicated in (a).\label{7}}
\end{figure*}

We first evaluated the method on a simple half-space model containing a low-resistivity anomaly, as shown in Fig. ~\ref{7}. The forward modeling results obtained with the conventional approach and the deep learning method are presented in Fig. ~\ref{7}. The results show that the deep learning method accurately predicts the overall forward response, demonstrating strong consistency and effectively capturing the characteristics of the low-resistivity anomaly. To enable a more detailed comparison, four points above the anomaly body were selected, as shown in Fig. ~\ref{7}p1-p4. The network-predicted responses closely match those computed with the conventional method.

To quantitatively assess the network’s predictions, we computed the relative errors between the network-predicted results and those obtained from conventional forward modeling, as shown in Table 2. The results show that the overall relative error is less than 2\%, indicating that the network can accurately reproduce the forward responses of the models.
\begin{table*}[h]
	\caption{Relative error between network-predicted forward responses and conventional forward modeling results for the simple model.}
	\begin{tabular}{cccccc}
		\tophline
		& All   & P1    & P2 & P3 & P4 \\
		\middlehline
	Relative error (\%) & 1.13    & 0.96 & 1.10& 0.89 & 1.03\\
		\bottomhline
	\end{tabular}
	\belowtable{} 
\end{table*}

\begin{figure*}[t]
	\includegraphics[width=18cm]{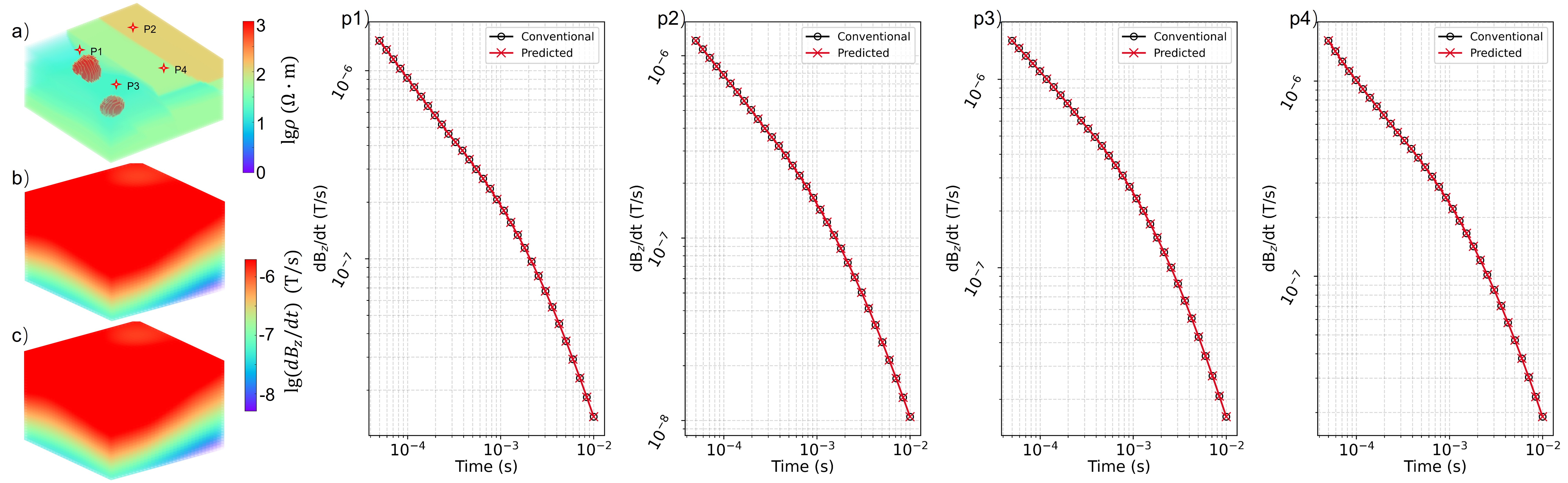}
	\caption{Complex model test: (a) geoelectric model, (b) network-predicted results, (c) forward modeling results using the conventional method, and (P1)–(P4) correspond to the responses at the four points indicated in (a).\label{8}}
\end{figure*}

To assess the network’s ability to handle complex models, we evaluated it on the most challenging model type. A randomly selected example is shown in Fig. ~\ref{8}a, which features faults, folds, and high-resistivity anomalies. The corresponding network-predicted results are presented in Fig. ~\ref{8}b. Even for such complex geological structures, the network accurately predicts the forward responses, demonstrating high reliability. For detailed comparison, four points were selected, as shown in Fig. ~\ref{8}p1–p4. The network-predicted responses closely match the numerical results from the conventional method, confirming the robustness of the network.

Table 3 summarizes the overall relative error for the complex models as well as the errors at the four specific comparison points. For complex models, the relative error increases slightly, with the maximum value of 1.42\% at point P2 and an overall error of 1.34\%, both remaining below 2\%. These results indicate that the network can still accurately predict the forward responses of complex geological models.
\begin{table*}[h]
	\caption{Relative error between network-predicted forward responses and conventional forward modeling results for the complex model.}
	\begin{tabular}{cccccc}
		\tophline
		& All   & P1    & P2 & P3 & P4 \\
		\middlehline
		Relative error (\%) & 1.34    & 1.19 & 1.42& 1.20 & 1.25\\
		\bottomhline
	\end{tabular}
	\belowtable{} 
\end{table*}

\begin{figure*}[t]
	\includegraphics[width=12cm]{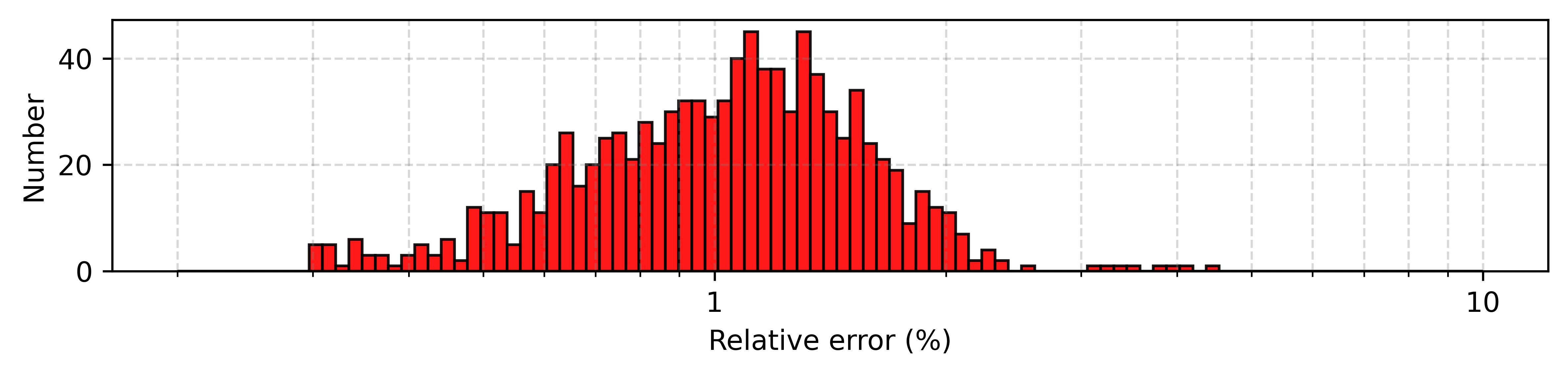}
	\caption{Relative error between network-predicted forward responses and conventional forward modeling results on the test set.\label{9}}
\end{figure*}

To further assess the reliability of the network, we calculated the relative errors between the network-predicted results and those obtained from conventional forward modeling across the entire test set, as shown in Fig. ~\ref{9}. The maximum relative error is 6.14\%, while 99\% of the models have relative errors below 5\%, with a mean of 1.21\%. These results demonstrate that the network effectively captures the mapping between geological models and their forward responses, making it suited for rapid forward modeling.

\section{Discussion}
As a large-scale, multi-structural dataset, OpenEM can substantially advance the application of deep learning in electromagnetic studies. Networks trained on OpenEM outperform those trained on random resistivity models or simple half-space models, as geological structures are inherently diverse and variable. OpenEM’s nine major structural model types cover the majority of geological scenarios from simple to complex, allowing networks trained on OpenEM to achieve more accurate and satisfactory results.

The resistivity range of OpenEM spans from 1 to 2000 Ω·m, encompassing most geological scenarios. In some formations, however, resistivity can exceed 2000 Ω·m, as observed in certain granites and basalts. Nevertheless, electromagnetic methods exhibit limited sensitivity and low resolution in such high-resistivity environments. Therefore, we set 2000 Ω·m as the upper resistivity limit. For requirements beyond this range, the provided 3D model generator enables customized model construction, allowing user-defined extensions to the overall model dataset.

3D forward modeling in electromagnetic methods is computationally intensive, which constrains the development of deep learning approaches for 3D applications. By employing the method developed by \cite{wang2026rapid}, reliable forward modeling can be achieved within a 5\% relative error. This margin of error is acceptable, as the uncertainty in measured field data is generally much higher than that introduced by forward modeling.

\section{Code and data availability}
The complete dataset is publicly available https://doi.org/10.5281/zenodo.17141981 \cite[]{OpenEM}. The 3D model generator is openly available at https://github.com/WAL-l/Geoelectrical3D.

\conclusions  
This paper introduces OpenEM, a large-scale, multi-structural 3D geoelectrical model dataset, aimed at promoting the development of deep learning methods for electromagnetic applications. OpenEM provides approximately 1.08 million geologically plausible 3D resistivity models with resistivity distributions ranging from 1 to 2000 Ω·m, and the structures include common geological features such as faults, folds, and various anomalies. Each model contains 3 to 7 layers, and in models with anomalies, 1 to 5 anomalies of both regular and irregular shapes are randomly distributed, thereby enhancing the dataset’s diversity and geological representativeness. To further enhance the usability of the dataset, OpenEM is accompanied by a fully controllable 3D model generator, enabling users to customize geological structures, resistivity distributions, fault geometries, folds, and anomalous bodies according to specific research requirements. This allows OpenEM to be easily extended beyond the predefined dataset and to adapt to application scenarios involving different geological environments or resistivity ranges. OpenEM is applicable to commonly used ground-based, airborne, semi-airborne, time-domain, and frequency-domain electromagnetic systems, aiming to provide a unified and comprehensive large-scale dataset to promote the application of deep learning methods in electromagnetics.

\authorcontribution{Conceptualization, methodology, software, validation, formal analysis, Shuang Wang.; data curation, Peifan Jiang.; writing---original draft preparation, Shuang Wang.; writing---review and editing, Jian Chen and Gianluca Fiandaca.; visualization, Fei Deng.; supervision, Xuben Wang.; project administration, Xuben Wang.} 

\competinginterests{The authors declare that they have no conflict of interest. } 

\begin{acknowledgements}
We are grateful to the developer of open-source Python packages SimPEG which greatly sped up our research work.  
\end{acknowledgements}

\financialsupport{This work was supported by the National Key Research and Development Program of China under Grant No. 2023YFB3905004. }



 \bibliographystyle{copernicus}
 \bibliography{r.bib}

\begin{thebibliography}{55}
\providecommand{\natexlab}[1]{#1}
\providecommand{\url}[1]{\texttt{#1}}
\providecommand{\urlprefix}{}
\expandafter\ifx\csname urlstyle\endcsname\relax
  \providecommand{\doi}[1]{https://doi.org/\discretionary{}{}{}#1}\else
  \providecommand{\doi}{https://doi.org/\discretionary{}{}{}\begingroup
  \urlstyle{rm}\Url}\fi

\bibitem[{Akingboye(2025)}]{akingboye2025electrical}
Akingboye, A.~S.: Electrical and seismic refraction methods: Fundamental
  concepts, current trends, and emerging machine learning prospects, Discover
  Geoscience, 3, 87, 2025.

\bibitem[{Asif et~al.(2021)Asif, Bording, Barfod, Grombacher, Maurya,
  Christiansen, Auken, and Larsen}]{asif2021effect}
Asif, M.~R., Bording, T.~S., Barfod, A.~S., Grombacher, D.~J., Maurya, P.~K.,
  Christiansen, A.~V., Auken, E., and Larsen, J.~J.: Effect of data
  pre-processing on the performance of neural networks for 1-D transient
  electromagnetic forward modeling, IEEE Access, 9, 34\,635--34\,646, 2021.

\bibitem[{Asif et~al.(2022)Asif, Maurya, Foged, Larsen, Auken, and
  Christiansen}]{asif2022automated}
Asif, M.~R., Maurya, P.~K., Foged, N., Larsen, J.~J., Auken, E., and
  Christiansen, A.~V.: Automated transient electromagnetic data processing for
  ground-based and airborne systems by a deep learning expert system, IEEE
  Transactions on Geoscience and Remote Sensing, 60, 1--14, 2022.

\bibitem[{Asif et~al.(2023)Asif, Foged, Bording, Larsen, and
  Christiansen}]{asif2023dl}
Asif, M.~R., Foged, N., Bording, T., Larsen, J.~J., and Christiansen, A.~V.:
  DL-RMD: a geophysically constrained electromagnetic resistivity model
  database (RMD) for deep learning (DL) applications, Earth System Science
  Data, 15, 1389--1401, 2023.

\bibitem[{Asif et~al.(2025)Asif, Kass, Herpe, Rawlinson, Westerhoff, Larsen,
  and Christiansen}]{asif2025comparative}
Asif, M.~R., Kass, M.~A., Herpe, M., Rawlinson, Z., Westerhoff, R., Larsen,
  J.~J., and Christiansen, A.~V.: Comparative analysis of deep learning and
  traditional airborne electromagnetic data processing: A case study,
  Geophysics, 90, WA103--WA112, 2025.

\bibitem[{Bai et~al.(2020)Bai, Vignoli, Viezzoli, Nevalainen, and
  Vacca}]{bai2020quasi}
Bai, P., Vignoli, G., Viezzoli, A., Nevalainen, J., and Vacca, G.: (Quasi-)
  real-time inversion of airborne time-domain electromagnetic data via
  artificial neural network, Remote Sensing, 12, 3440, 2020.

\bibitem[{Ball et~al.(2020)Ball, Bedrosian, and Minsley}]{ball2020high}
Ball, L.~B., Bedrosian, P.~A., and Minsley, B.~J.: High-resolution mapping of
  the freshwater--brine interface using deterministic and Bayesian inversion of
  airborne electromagnetic data at Paradox Valley, USA, Hydrogeology Journal,
  28, 941--954, 2020.

\bibitem[{Bedrosian et~al.(2014)Bedrosian, Ball, and
  Bloss}]{bedrosian2014airborne}
Bedrosian, P.~A., Ball, L.~B., and Bloss, B.~R.: Airborne Electromagnetic Data
  and Processing Within Leech Lake Basin, Fort Irwin, California, US Department
  of the Interior, US Geological Survey, 2014.

\bibitem[{Bergen et~al.(2019)Bergen, Johnson, de~Hoop, and
  Beroza}]{bergen2019machine}
Bergen, K.~J., Johnson, P.~A., de~Hoop, M.~V., and Beroza, G.~C.: Machine
  learning for data-driven discovery in solid Earth geoscience, Science, 363,
  eaau0323, 2019.

\bibitem[{Bording et~al.(2021)Bording, Asif, Barfod, Larsen, Zhang, Grombacher,
  Christiansen, Engebretsen, Pedersen, Maurya et~al.}]{bording2021machine}
Bording, T.~S., Asif, M.~R., Barfod, A.~S., Larsen, J.~J., Zhang, B.,
  Grombacher, D.~J., Christiansen, A.~V., Engebretsen, K.~W., Pedersen, J.~B.,
  Maurya, P.~K., et~al.: Machine learning based fast forward modelling of
  ground-based time-domain electromagnetic data, Journal of Applied Geophysics,
  187, 104\,290, 2021.

\bibitem[{Chen et~al.(2022)Chen, Zhang, and Lin}]{chen2022transient}
Chen, J., Zhang, Y., and Lin, T.: Transient electromagnetic machine learning
  inversion based on pseudo wave field data, IEEE Transactions on Geoscience
  and Remote Sensing, 60, 1--10, 2022.

\bibitem[{Chen et~al.(2025)Chen, Galli, Signora, Sullivan, Zhang, and
  Fiandaca}]{chen2025rapid}
Chen, J., Galli, S., Signora, A., Sullivan, N. A.~L., Zhang, B., and Fiandaca,
  G.: Rapid Bayesian Imaging of Large-Scale Transient Electromagnetic Data
  Using Probabilistic Neural Networks, Journal of Geophysical Research: Machine
  Learning and Computation, 2, e2024JH000\,536, 2025.

\bibitem[{Christensen et~al.(2017)Christensen, Ferre, Fiandaca, and
  Christensen}]{christensen2017voxel}
Christensen, N.~K., Ferre, T. P.~A., Fiandaca, G., and Christensen, S.: Voxel
  inversion of airborne electromagnetic data for improved groundwater model
  construction and prediction accuracy, Hydrology and Earth System Sciences,
  21, 1321--1337, 2017.

\bibitem[{Cockett et~al.(2015)Cockett, Kang, Heagy, Pidlisecky, and
  Oldenburg}]{cockett2015simpeg}
Cockett, R., Kang, S., Heagy, L.~J., Pidlisecky, A., and Oldenburg, D.~W.:
  SimPEG: An open source framework for simulation and gradient based parameter
  estimation in geophysical applications, Computers \& Geosciences, 85,
  142--154, 2015.

\bibitem[{Damhuis et~al.(2020)Damhuis, Roux, and
  Fourie}]{damhuis2020identification}
Damhuis, R.~M., Roux, P.~L., and Fourie, C.~J.: The identification and
  mitigation of geohazards using shallow airborne engineering geophysics and
  land-based geophysics for brown-and greenfield road investigations, Quarterly
  Journal of Engineering Geology and Hydrogeology, 53, 321--332, 2020.

\bibitem[{Dzikunoo et~al.(2020)Dzikunoo, Vignoli, J{\o}rgensen, Yidana, and
  Banoeng-Yakubo}]{dzikunoo2020new}
Dzikunoo, E.~A., Vignoli, G., J{\o}rgensen, F., Yidana, S.~M., and
  Banoeng-Yakubo, B.: New regional stratigraphic insights from a 3D geological
  model of the Nasia sub-basin, Ghana, developed for hydrogeological purposes
  and based on reprocessed B-field data originally collected for mineral
  exploration, Solid Earth, 11, 349--361, 2020.

\bibitem[{Huang et~al.(2025)Huang, Wu, and Xue}]{huang2025data}
Huang, Q., Wu, S., and Xue, J.: Data Science and Machine Learning in
  Geo-Electromagnetics: A Review, Surveys in Geophysics, pp. 1--56, 2025.

\bibitem[{J{\o}rgensen et~al.(2005)J{\o}rgensen, Sandersen, Auken,
  Lykke-Andersen, and S{\o}rensen}]{jorgensen2005contributions}
J{\o}rgensen, F., Sandersen, P.~B., Auken, E., Lykke-Andersen, H., and
  S{\o}rensen, K.: Contributions to the geological mapping of Mors, Denmark--a
  study based on a large-scale TEM survey, Bulletin of the Geological Society
  of Denmark, 52, 53--75, 2005.

\bibitem[{Kon{\'e} et~al.(2021)Kon{\'e}, Nasr, Traor{\'e}, Amiri, Inoubli,
  Sangar{\'e}, and Qaysi}]{kone2021geophysical}
Kon{\'e}, A.~Y., Nasr, I.~H., Traor{\'e}, B., Amiri, A., Inoubli, M.~H.,
  Sangar{\'e}, S., and Qaysi, S.: Geophysical contributions to gold exploration
  in western Mali according to airborne electromagnetic data interpretations,
  Minerals, 11, 126, 2021.

\bibitem[{Li et~al.(2024)Li, Wu, Cai, Chen, Chen, Xiao, and Yan}]{li2024gtcn}
Li, G., Wu, S., Cai, H., Chen, C., Chen, H., Xiao, D., and Yan, J.: GTCN: Gated
  Temporal Convolutional Networks for Controlled-Source Electromagnetic Data
  Denoising, IEEE Transactions on Geoscience and Remote Sensing, 2024.

\bibitem[{Liu et~al.(2024)Liu, Zhang, Guo, Zhang, Kang, and
  Zhao}]{liu2024multi}
Liu, Y., Zhang, Y., Guo, C., Zhang, S., Kang, H., and Zhao, Q.: A multi-task
  learning network based on the Transformer network for airborne
  electromagnetic detection imaging and denoising, Journal of Geophysics and
  Engineering, 21, 1056--1070, 2024.

\bibitem[{Malehmir et~al.(2016)Malehmir, Socco, Bastani, Krawczyk, Pfaffhuber,
  Miller, Maurer, Frauenfelder, Suto, Bazin et~al.}]{malehmir2016near}
Malehmir, A., Socco, L.~V., Bastani, M., Krawczyk, C.~M., Pfaffhuber, A.~A.,
  Miller, R.~D., Maurer, H., Frauenfelder, R., Suto, K., Bazin, S., et~al.:
  Near-surface geophysical characterization of areas prone to natural hazards:
  a review of the current and perspective on the future, Advances in
  Geophysics, 57, 51--146, 2016.

\bibitem[{Melo(2018)}]{melo2018integrated}
Melo, A.~T.: Integrated quantitative interpretation of multiple geophysical
  data for geology differentiation, Colorado School of Mines, 2018.

\bibitem[{Minsley et~al.(2021)Minsley, Rigby, James, Burton, Knierim, Pace,
  Bedrosian, and Kress}]{minsley2021airborne}
Minsley, B.~J., Rigby, J.~R., James, S.~R., Burton, B.~L., Knierim, K.~J.,
  Pace, M.~D., Bedrosian, P.~A., and Kress, W.~H.: Airborne geophysical surveys
  of the lower Mississippi Valley demonstrate system-scale mapping of
  subsurface architecture, Communications Earth \& Environment, 2, 131, 2021.

\bibitem[{M{\o}ller et~al.(2001)M{\o}ller, Jacobsen, and
  Christensen}]{moller2001rapid}
M{\o}ller, I., Jacobsen, B.~H., and Christensen, N.~B.: Rapid inversion of 2-D
  geoelectrical data by multichannel deconvolution, Geophysics, 66, 800--808,
  2001.

\bibitem[{Okada(2021)}]{okada2021historical}
Okada, K.: A historical overview of the past three decades of mineral
  exploration technology, Natural Resources Research, 30, 2839--2860, 2021.

\bibitem[{Oldenburg et~al.(2020)Oldenburg, Heagy, Kang, and
  Cockett}]{oldenburg20203d}
Oldenburg, D.~W., Heagy, L.~J., Kang, S., and Cockett, R.: 3D electromagnetic
  modelling and inversion: a case for open source, Exploration Geophysics, 51,
  25--37, 2020.

\bibitem[{Qu et~al.(2025)Qu, Gao, Xing, and Zhang}]{qu2025deep}
Qu, Z., Gao, Y., Xing, K., and Zhang, X.: Deep-TEMNet: A Hybrid U-Net--2D LSTM
  Network for Efficient and Accurate 2.5 D Transient Electromagnetic Forward
  Modeling, Remote Sensing, 17, 264, 2025.

\bibitem[{Reichstein et~al.(2019)Reichstein, Camps-Valls, Stevens, Jung,
  Denzler, Carvalhais, and Prabhat}]{reichstein2019deep}
Reichstein, M., Camps-Valls, G., Stevens, B., Jung, M., Denzler, J.,
  Carvalhais, N., and Prabhat, F.: Deep learning and process understanding for
  data-driven Earth system science, Nature, 566, 195--204, 2019.

\bibitem[{Siemon et~al.(2009)Siemon, Auken, and
  Christiansen}]{siemon2009laterally}
Siemon, B., Auken, E., and Christiansen, A.~V.: Laterally constrained inversion
  of helicopter-borne frequency-domain electromagnetic data, Journal of Applied
  Geophysics, 67, 259--268, 2009.

\bibitem[{Tang et~al.(2024)Tang, Gan, Li, and Shen}]{tang2024fast}
Tang, R., Gan, L., Li, F., and Shen, F.: A Fast Three-dimensional Imaging
  Scheme of Airborne Time Domain Electromagnetic Data using Deep Learning,
  Authorea Preprints, 2024.

\bibitem[{Teklesenbet(2012)}]{teklesenbet2012multidimensional}
Teklesenbet, A.: Multidimensional inversion of MT data from Alid Geothermal
  area, Eritrea. Comparison with geological structures and identification of a
  geothermal reservoir, Ph.D. thesis, 2012.

\bibitem[{Vall{\'e}e and Smith(2009)}]{vallee2009inversion}
Vall{\'e}e, M.~A. and Smith, R.~S.: Inversion of airborne time-domain
  electromagnetic data to a 1D structure using lateral constraints, Near
  Surface Geophysics, 7, 63--71, 2009.

\bibitem[{Vignoli et~al.(2015)Vignoli, Fiandaca, Christiansen, Kirkegaard, and
  Auken}]{vignoli2015sharp}
Vignoli, G., Fiandaca, G., Christiansen, A.~V., Kirkegaard, C., and Auken, E.:
  Sharp spatially constrained inversion with applications to transient
  electromagnetic data, Geophysical Prospecting, 63, 243--255, 2015.

\bibitem[{Wang et~al.(2025{\natexlab{a}})Wang, Wang, and
  Deng}]{wang2025interpretable}
Wang, S., Wang, X., and Deng, F.: Interpretable deep learning paradigm for
  airborne transient electromagnetic inversion, in: SEG Near-Surface
  Geophysical Exploration and Geo-Disaster Prevention Technology Workshop,
  Chengdu, China, July 4--6, 2025, pp. 51--52, Society of Exploration
  Geophysicists, 2025{\natexlab{a}}.

\bibitem[{Wang et~al.(2025{\natexlab{b}})Wang, Wang, Deng, Jiang, Chen, and
  Gianluca}]{OpenEM}
Wang, S., Wang, X., Deng, F., Jiang, P., Chen, J., and Gianluca, F.: OpenEM:
  Large-scale multi-structural 3D datasets for electromagnetic methods, Zenodo,
  https://doi.org/10.5281/zenodo.17141981, 2025{\natexlab{b}}.

\bibitem[{Wang et~al.(2026{\natexlab{a}})Wang, Guo, Wang, Deng, Mao, Wang, and
  Gao}]{wang2025dremnet}
Wang, S., Guo, M., Wang, X., Deng, F., Mao, L., Wang, B., and Gao, W.: DREMnet:
  An interpretable denoising framework for semi-airborne transient
  electromagnetic signal, IEEE Transactions on Geoscience and Remote Sensing,
  64, 1--12, 2026{\natexlab{a}}.

\bibitem[{Wang et~al.(2026{\natexlab{b}})Wang, Wang, Jiang, Deng, and
  Li}]{wang2026rapid}
Wang, X., Wang, S., Jiang, P., Deng, F., and Li, Y.: Rapid and high-accuracy
  three-dimensional airborne transient electromagnetic forward modeling based
  on machine learning, Journal of Geophysical Research: Machine Learning and
  Computation, 3, e2025JH001\,181, 2026{\natexlab{b}}.

\bibitem[{Wong et~al.(2020)Wong, Ley-Cooper, Rollet, Brodie, Bonnardot,
  English, Nicoll, and Roach}]{wong2020interpretation}
Wong, S.~C., Ley-Cooper, Y., Rollet, N., Brodie, R.~C., Bonnardot, M.-A.,
  English, P., Nicoll, M., and Roach, I.: Interpretation of the AusAEM1:
  Insights from the world's largest airborne electromagnetic survey, Geoscience
  Australia, 2020.

\bibitem[{Wu et~al.(2021{\natexlab{a}})Wu, Huang, and
  Zhao}]{wu2021convolutional}
Wu, S., Huang, Q., and Zhao, L.: Convolutional neural network inversion of
  airborne transient electromagnetic data, Geophysical Prospecting, 69,
  1761--1772, 2021{\natexlab{a}}.

\bibitem[{Wu et~al.(2021{\natexlab{b}})Wu, Huang, and Zhao}]{wu2021noising}
Wu, S., Huang, Q., and Zhao, L.: De-noising of transient electromagnetic data
  based on the long short-term memory-autoencoder, Geophysical Journal
  International, 224, 669--681, 2021{\natexlab{b}}.

\bibitem[{Wu et~al.(2022{\natexlab{a}})Wu, Huang, and
  Zhao}]{wu2022instantaneous}
Wu, S., Huang, Q., and Zhao, L.: Instantaneous inversion of airborne
  electromagnetic data based on deep learning, Geophysical Research Letters,
  49, e2021GL097\,165, 2022{\natexlab{a}}.

\bibitem[{Wu et~al.(2023{\natexlab{a}})Wu, Huang, and Zhao}]{wu2023deep}
Wu, S., Huang, Q., and Zhao, L.: A deep learning-based network for the
  simulation of airborne electromagnetic responses, Geophysical Journal
  International, 233, 253--263, 2023{\natexlab{a}}.

\bibitem[{Wu et~al.(2023{\natexlab{b}})Wu, Huang, and Zhao}]{wu2023fast}
Wu, S., Huang, Q., and Zhao, L.: Fast Bayesian inversion of airborne
  electromagnetic data based on the invertible neural network, IEEE
  Transactions on Geoscience and Remote Sensing, 61, 1--11, 2023{\natexlab{b}}.

\bibitem[{Wu et~al.(2024)Wu, Huang, and Zhao}]{wu2024physics}
Wu, S., Huang, Q., and Zhao, L.: Physics-guided deep learning-based inversion
  for airborne electromagnetic data, Geophysical Journal International, 238,
  1774--1789, 2024.

\bibitem[{Wu et~al.(2025{\natexlab{a}})Wu, Deng, Wei, Rizwan~Asif, Vignoli, and
  Farquharson}]{wu2025frontiers}
Wu, S., Deng, S., Wei, X., Rizwan~Asif, M., Vignoli, G., and Farquharson,
  C.~G.: Frontiers in electromagnetic geophysics—Introduction,
  2025{\natexlab{a}}.

\bibitem[{Wu et~al.(2025{\natexlab{b}})Wu, Sun, and Chen}]{wu2025variational}
Wu, S., Sun, J., and Chen, J.: Variational inference for geophysical Bayesian
  inverse problems using normalizing flows: An unsupervised approach to
  electromagnetic data inversion, Geophysical Journal International, p.
  ggaf239, 2025{\natexlab{b}}.

\bibitem[{Wu et~al.(2025{\natexlab{c}})Wu, Thoram, Sun, Sager, and
  Chen}]{wu2025characterizing}
Wu, S., Thoram, S., Sun, J., Sager, W., and Chen, J.: Characterizing marine
  magnetic anomalies: A machine learning approach to advancing the
  understanding of oceanic crust formation, Journal of Geophysical Research:
  Solid Earth, 130, e2024JB030\,682, 2025{\natexlab{c}}.

\bibitem[{Wu et~al.(2019)Wu, Xue, Xiao, Li, Liu, and Fang}]{wu2019removal}
Wu, X., Xue, G., Xiao, P., Li, J., Liu, L., and Fang, G.: The removal of the
  high-frequency motion-induced noise in helicopter-borne transient
  electromagnetic data based on wavelet neural network, Geophysics, 84, K1--K9,
  2019.

\bibitem[{Wu et~al.(2020)Wu, Xue, He, and Xue}]{wu2020removal}
Wu, X., Xue, G., He, Y., and Xue, J.: Removal of multisource noise in airborne
  electromagnetic data based on deep learning, Geophysics, 85, B207--B222,
  2020.

\bibitem[{Wu et~al.(2022{\natexlab{b}})Wu, Xue, Zhao, Lv, Zhou, and
  Shi}]{wu2022deep}
Wu, X., Xue, G., Zhao, Y., Lv, P., Zhou, Z., and Shi, J.: A deep learning
  estimation of the earth resistivity model for the airborne transient
  electromagnetic observation, Journal of Geophysical Research: Solid Earth,
  127, e2021JB023\,185, 2022{\natexlab{b}}.

\bibitem[{Yang and Oldenburg(2012)}]{yang2012three}
Yang, D. and Oldenburg, D.~W.: Three-dimensional inversion of airborne
  time-domain electromagnetic data with applications to a porphyry deposit,
  Geophysics, 77, B23--B34, 2012.

\bibitem[{Zhang et~al.(2025)Zhang, Liu, Feng, Wang, Liu, Luo, Zheng, and
  Sun}]{zhang20253d}
Zhang, S., Liu, R., Feng, H., Wang, Z., Liu, S., Luo, Y., Zheng, Z., and Sun,
  H.: 3D inversion of airborne transient electromagnetic data using deep
  learning, Journal of Applied Geophysics, 239, 105\,737, 2025.

\bibitem[{Zhao et~al.(2024)Zhao, Wu, Chen, Xue, and Shi}]{zhao2024three}
Zhao, Y., Wu, X., Chen, W., Xue, J., and Shi, J.: Three-dimensional inversion
  for short-offset transient electromagnetic data based on 3D U-Net, Journal of
  Geophysics and Engineering, 21, 922--937, 2024.

\bibitem[{Zhu et~al.(2025)Zhu, Tang, Li, Hu, and Peng}]{zhu2025deep}
Zhu, Y., Tang, Y., Li, J., Hu, X., and Peng, R.: A Deep Learning Approach for
  Transient Electromagnetic Data Denoising, Inversion and Uncertainty Analysis
  With Monte Carlo Dropout Technique, Geophysical Prospecting, 73, e70\,069,
  2025.

\end{thebibliography}

\end{document}